\documentclass[preprint,review,12pt,numbers]{elsarticle}

\usepackage[utf8]{inputenc}
\usepackage[T1]{fontenc}
\usepackage{amsmath,amssymb,amsfonts}
\usepackage{booktabs}
\usepackage{graphicx}
\usepackage{hyperref}
\usepackage{xcolor}
\usepackage{multirow}
\usepackage{algorithm}
\usepackage{algorithmic}
\usepackage{subcaption}
\usepackage{microtype}
\usepackage{float}
\graphicspath{{figures/}}

\hypersetup{
    colorlinks=true,
    linkcolor=blue!60!black,
    citecolor=blue!60!black,
    urlcolor=blue!60!black
}

\begin{document}

\begin{frontmatter}

\title{How Pruning Reshapes Features: Sparse Autoencoder Analysis of Weight-Pruned Language Models}

\author[inst1]{Hector Borobia\fnref{cor1}}
\ead{hecboar@doctor.upv.es}
\author[inst2]{Elies Segu\'i-Mas}
\ead{eseguim@upvnet.upv.es}
\author[inst3]{Guillermina Tormo-Carb\'o}
\ead{gtormo@omp.upv.es}

\fntext[cor1]{Corresponding author.}

\affiliation[inst1]{organization={VRAIN -- Valencian Research Institute for Artificial Intelligence, Universitat Polit\`ecnica de Val\`encia},
    city={Valencia},
    country={Spain}}

\affiliation[inst2]{organization={Department of Economics and Social Sciences, Universitat Polit\`ecnica de Val\`encia},
    city={Valencia},
    country={Spain}}

\affiliation[inst3]{organization={Department of Business Organisation, Universitat Polit\`ecnica de Val\`encia},
    city={Valencia},
    country={Spain}}

\begin{abstract}
Weight pruning is a standard technique for compressing large language models, yet its effect on learned internal representations remains poorly understood. We present the first systematic study of how unstructured pruning reshapes the feature geometry of language models, using Sparse Autoencoders (SAEs) as interpretability probes. Across three model families (Gemma~3 1B, Gemma~2 2B, Llama~3.2 1B), two pruning methods (magnitude and Wanda), and six sparsity levels (0--60\%), we investigate five research questions spanning seed stability, feature survival, SAE transferability, feature fragility, and causal relevance. Our most striking finding is that \emph{rare} SAE features---those with low firing rates---survive pruning far better than frequent ones, with within-condition Spearman correlations of $\rho = -1.0$ in 11 of 17 experimental conditions. This counter-intuitive result suggests that pruning acts as implicit feature selection, preferentially destroying high-frequency generic features while preserving specialized rare ones. We further show that Wanda pruning preserves feature structure up to $3.7\times$ better than magnitude pruning, that pre-trained SAEs remain viable on Wanda-pruned models up to 50\% sparsity, and that geometric feature survival does not predict causal importance---a dissociation with implications for interpretability under compression.
\end{abstract}


\end{frontmatter}

\section{Introduction}
\label{sec:intro}

Weight pruning has become a standard approach for deploying large language models (LLMs) under computational constraints~\cite{frantar2023sparsegpt,sun2024wanda}. By setting a fraction of model weights to zero, practitioners can reduce memory footprint and, with appropriate hardware or software support, accelerate inference. A substantial body of work has characterized the impact of pruning on task-level metrics such as perplexity and downstream accuracy. However, a fundamental question remains largely unexplored: \emph{what happens to the internal features of a language model when it is pruned?}

Recent advances in mechanistic interpretability have provided tools for answering this question. Sparse Autoencoders (SAEs)~\cite{bricken2023monosemanticity,cunningham2023sparse} decompose a model's dense activation vectors into sparse, interpretable features. By training SAEs on activations from both dense and pruned models, we can directly compare the feature dictionaries that each model has learned, measuring which features survive, which are destroyed, and which emerge anew.

In this work, we present the first systematic study of how unstructured weight pruning reshapes the internal feature geometry of LLMs as revealed by SAEs. We investigate five research questions:

\begin{enumerate}
    \item[\textbf{RQ1}] \textbf{Seed Stability:} Are SAE features reproducible across random seeds, and does pruning affect this reproducibility?
    \item[\textbf{RQ2}] \textbf{Feature Survival:} When SAEs are trained on dense versus pruned models, how many features can be matched across the two dictionaries?
    \item[\textbf{RQ3}] \textbf{Transferability:} Can a pre-trained SAE designed for the dense model still reconstruct pruned activations?
    \item[\textbf{RQ4}] \textbf{Fragility Taxonomy:} Are rare features (low firing rate) more or less likely to survive pruning than frequent ones?
    \item[\textbf{RQ5}] \textbf{Causal Relevance:} Do geometrically surviving features have greater causal effect on model outputs than those that are destroyed?
\end{enumerate}

Our experiments span three model families (Gemma~3 1B, Gemma~2 2B, Llama~3.2 1B), two pruning methods (global magnitude pruning and Wanda~\cite{sun2024wanda}), and six sparsity levels from 0\% to 60\%, totaling 22 successful experimental runs with 3--5 SAE seeds each.

Our most striking finding is counter-intuitive: \emph{rare} SAE features survive pruning far better than frequent ones. Across all three model families, the within-condition Spearman correlation between firing rate and survival rate is $\rho = -1.0$ in 11 of 17 conditions, and $\rho \leq -0.8$ in 15 of 17, with the remaining two including edge cases where catastrophic pruning destroys nearly all features regardless of firing rate. For example, in Gemma~3 at 30\% magnitude sparsity, the rarest features (Q1 quintile) survive at a rate of 76.0\%, while the most frequent ones (Q5) survive at only 14.6\%---a $5.2\times$ differential. This pattern is consistent across all models and both pruning methods, suggesting that pruning performs implicit feature selection by preferentially destroying generic, high-frequency features while preserving specialized information.

We additionally show that (i) Wanda preserves feature structure up to $3.7\times$ better than magnitude pruning at matched sparsity; (ii) pre-trained (official) SAEs remain viable on Wanda-pruned activations up to 50\% sparsity; (iii) seed stability is low in absolute terms (MNN $\approx 2$--4\% at $\tau=0.7$) but the \emph{pattern} of degradation is highly consistent; and (iv) geometric feature survival does not predict causal importance---robust and fragile features contribute equally to model outputs.

\section{Related Work}
\label{sec:related}

\paragraph{Sparse Autoencoders for Interpretability.}
Sparse Autoencoders have emerged as a leading tool for extracting interpretable features from neural network activations. Bricken et al.~\cite{bricken2023monosemanticity} demonstrated that SAEs can decompose MLP activations in language models into monosemantic features. Cunningham et al.~\cite{cunningham2023sparse} proposed using SAEs to identify features across different training runs and architectures. The Gemma Scope project~\cite{lieberum2024gemma} released pre-trained SAEs for the Gemma model family, providing standardized reference dictionaries. Templeton et al.~\cite{templeton2024scaling} scaled SAE training to large models, showing that interpretable features persist at scale. Recent work has explored TopK activation functions~\cite{makhzani2014k} and improved training procedures to reduce dead features. Our work builds on these foundations by asking whether the features SAEs discover are stable under a perturbation that practitioners commonly apply: weight pruning.

\paragraph{Weight Pruning.}
Unstructured weight pruning removes individual weights from neural networks. Classical approaches use weight magnitude as a pruning criterion~\cite{han2015learning}. Frantar and Alistarh~\cite{frantar2023sparsegpt} introduced SparseGPT, enabling one-shot pruning of LLMs without retraining. Sun et al.~\cite{sun2024wanda} proposed Wanda, which prunes weights based on the product of weight magnitude and input activation norm, achieving competitive results with minimal calibration data. The Lottery Ticket Hypothesis~\cite{frankle2019lottery} posits that sparse subnetworks within dense models can match the performance of the full model. Our work connects these two lines of research by examining what pruning does to the \emph{features} of a model, rather than only its task performance.

\paragraph{Intersection: Interpretability Under Compression.}
Despite extensive work on both SAEs and pruning independently, their intersection remains largely unexplored. Prior work has examined how quantization affects model representations~\cite{dettmers2024qlora} and how knowledge distillation preserves or alters learned features. To our knowledge, no prior work has systematically studied how weight pruning affects the SAE feature dictionary of a language model across multiple model families, pruning methods, and sparsity levels. This gap is significant because practitioners who prune models for deployment may also wish to use interpretability tools---and it is unclear whether interpretability artifacts trained on the dense model remain valid after pruning.

\section{Method}
\label{sec:method}

\subsection{Models and Pruning}
\label{sec:models}

We study three pre-trained language models spanning two architectural families: \textbf{Gemma~3 1B}~\cite{team2024gemma} (primary model, most extensive sparsity grid), \textbf{Gemma~2 2B}~\cite{team2024gemma2} (anchor for replication with available official SAEs), and \textbf{Llama~3.2 1B}~\cite{touvron2023llama} (cross-family validation). For each model, we extract activations from the residual stream at a mid-layer position: layer~12 for both Gemma models and layer~8 for Llama.

We apply two unstructured pruning methods:
\begin{itemize}
    \item \textbf{Magnitude pruning}: A global threshold is computed over all linear layer weights, and weights below this threshold are zeroed. This is the simplest and most widely used pruning criterion.
    \item \textbf{Wanda}~\cite{sun2024wanda}: For each row of each weight matrix, the pruning score is the element-wise product of weight magnitude and input activation norm (computed from 300K calibration tokens from FineWeb-Edu~\cite{penedo2024fineweb}). This activation-aware criterion preserves weights that are both large and frequently activated.
\end{itemize}

We evaluate sparsity levels $s \in \{0.0, 0.2, 0.3, 0.4, 0.5, 0.6\}$ for Gemma~3 (the primary model), and a subset for Gemma~2 and Llama. The full experimental matrix comprises 24 planned runs, of which 22 completed successfully (two runs at layer~18 of Gemma~3 produced NaN activations due to float16 overflow at deep layers). Table~\ref{tab:model_quality} reports perplexity on WikiText-2 for all configurations.

\subsection{SAE Architecture and Training}
\label{sec:sae}

We train TopK Sparse Autoencoders~\cite{makhzani2014k} on residual stream activations. Given input $\mathbf{x} \in \mathbb{R}^{d}$, the SAE computes:
\begin{align}
    \mathbf{z} &= \text{TopK}\left(\mathbf{W}_{\text{enc}} \hat{\mathbf{x}} + \mathbf{b}_{\text{enc}},\; k\right), \\
    \hat{\mathbf{x}}_{\text{rec}} &= \mathbf{W}_{\text{dec}} \mathbf{z} + \mathbf{b}_{\text{dec}},
\end{align}
where $\hat{\mathbf{x}} = (\mathbf{x} - \boldsymbol{\mu}) / \boldsymbol{\sigma}$ is the input normalized to zero mean and unit variance (statistics computed in float64 from 50K activation samples), and $\text{TopK}(\cdot, k)$ retains only the $k$ largest activations, setting the rest to zero.

Key hyperparameters: $d_{\text{SAE}} = 8 \times d_{\text{model}}$ (expansion factor), $k = 64$, learning rate $5 \times 10^{-5}$ with cosine annealing, 20{,}000 training steps, batch size 2{,}048--4{,}096 (GPU-dependent), and dead feature resampling every 2{,}000 steps. Training activations are extracted from 1.5M tokens of FineWeb-Edu, and evaluation uses a held-out set of 300K tokens.

For each experimental condition, we train 3--5 SAEs with different random seeds. This enables seed stability analysis (RQ1) and provides multiple reference points for feature matching.

\subsection{Feature Matching Metrics}
\label{sec:matching}

To compare SAE feature dictionaries, we extract the decoder weight matrix $\mathbf{W}_{\text{dec}} \in \mathbb{R}^{d_{\text{SAE}} \times d}$ and $\ell_2$-normalize each column to obtain unit-norm feature vectors. We then compute pairwise cosine similarities between feature vectors from two SAEs (A and B) and apply three matching metrics at threshold $\tau$:

\begin{itemize}
    \item \textbf{One-way match rate}: The fraction of features in A whose best match in B has cosine similarity $\geq \tau$.
    \item \textbf{Mutual Nearest Neighbor (MNN) rate}: The fraction of features in A that are mutual nearest neighbors with their best match in B (both directions), with similarity $\geq \tau$.
    \item \textbf{Greedy 1-to-1 matching}: Starting from the highest-similarity pairs and proceeding greedily, the fraction of features in A that can be uniquely matched to a feature in B with similarity $\geq \tau$.
\end{itemize}

We report results at $\tau \in \{0.5, 0.6, 0.7, 0.8, 0.9\}$, with $\tau = 0.7$ as the primary threshold. We use MNN as our primary metric because it is the most conservative: it requires that features be best matches in \emph{both} directions.

\subsection{Causal Ablation Protocol}
\label{sec:causal}

To assess whether geometrically surviving features are causally more important than fragile ones (RQ5), we perform single-feature ablation experiments. For a given dense--pruned pair, we classify features as ``robust'' (top 100 by cosine match score) or ``fragile'' (bottom 100). For each feature $f_i$, we:
\begin{enumerate}
    \item Encode the residual stream activation through the SAE.
    \item Zero out the activation of feature $f_i$.
    \item Decode back and inject the modified activation into the model via a forward hook at the target layer.
    \item Measure the KL divergence between the original and ablated output distributions at the final token position.
\end{enumerate}
We average over 64 prompts (256 tokens each) from FineWeb-Edu and report the mean KL divergence per category.

\section{Results}
\label{sec:results}

\subsection{Model Quality Under Pruning}
\label{sec:model_quality}

Table~\ref{tab:model_quality} reports WikiText-2 perplexity for all model--method--sparsity combinations. Llama~3.2 1B exhibits the most graceful degradation: Wanda at 30\% sparsity increases perplexity from 17.7 to only 20.9 ($1.2\times$), and even at 50\% sparsity, Wanda achieves a perplexity of 53.1 compared to magnitude pruning's 2{,}583. Gemma~3 1B shows steeper degradation, with magnitude pruning becoming catastrophic ($>10^6$) at $\geq$50\% sparsity. Across all models, Wanda consistently outperforms magnitude pruning at matched sparsity.

\begin{table}[t]
\centering
\caption{WikiText-2 perplexity across models, pruning methods, and sparsity levels. Lower is better. Gemma~2 dense perplexity is anomalously high (likely due to context length mismatch); relative differences remain informative. Catastrophic values ($>10^5$) are abbreviated.}
\label{tab:model_quality}
\small
\begin{tabular}{llrrrrrr}
\toprule
\textbf{Model} & \textbf{Method} & \textbf{0\%} & \textbf{20\%} & \textbf{30\%} & \textbf{40\%} & \textbf{50\%} & \textbf{60\%} \\
\midrule
\multirow{2}{*}{Gemma 3 1B} & Magnitude & 64.2 & 93.8 & 333 & 4{,}472 & $>$10$^6$ & $>$10$^6$ \\
                             & Wanda     & --- & 286 & 822 & 9{,}315 & 34{,}020 & 110{,}751 \\
\midrule
\multirow{2}{*}{Gemma 2 2B} & Magnitude & 410 & --- & 2{,}359 & --- & 303{,}290 & --- \\
                             & Wanda     & --- & --- & 489 & --- & 904 & --- \\
\midrule
\multirow{2}{*}{Llama 3.2 1B} & Magnitude & 17.7 & --- & --- & --- & 2{,}583 & --- \\
                               & Wanda     & --- & --- & 20.9 & --- & 53.1 & --- \\
\bottomrule
\end{tabular}
\end{table}

The large majority of SAEs achieved near-complete feature utilization across all 80 trained models, validating our training procedure (learning rate $5 \times 10^{-5}$, dead feature resampling every 2{,}000 steps). SAE reconstruction quality (FVU) showed an interesting divergence across model families: in Gemma~3, FVU \emph{decreased} from 0.367 (dense) to 0.241 at 60\% Wanda sparsity, suggesting simpler activation geometry in pruned models. In contrast, Llama FVU \emph{increased} from 0.356 (dense) to 0.487 at 50\% magnitude sparsity, indicating more complex or chaotic activations. Gemma~2 showed a non-monotonic pattern ($0.409 \to 0.377 \to 0.439$).

\subsection{RQ1: Seed Stability}
\label{sec:rq1}

\begin{figure}[t]
    \centering
    \includegraphics[width=\textwidth]{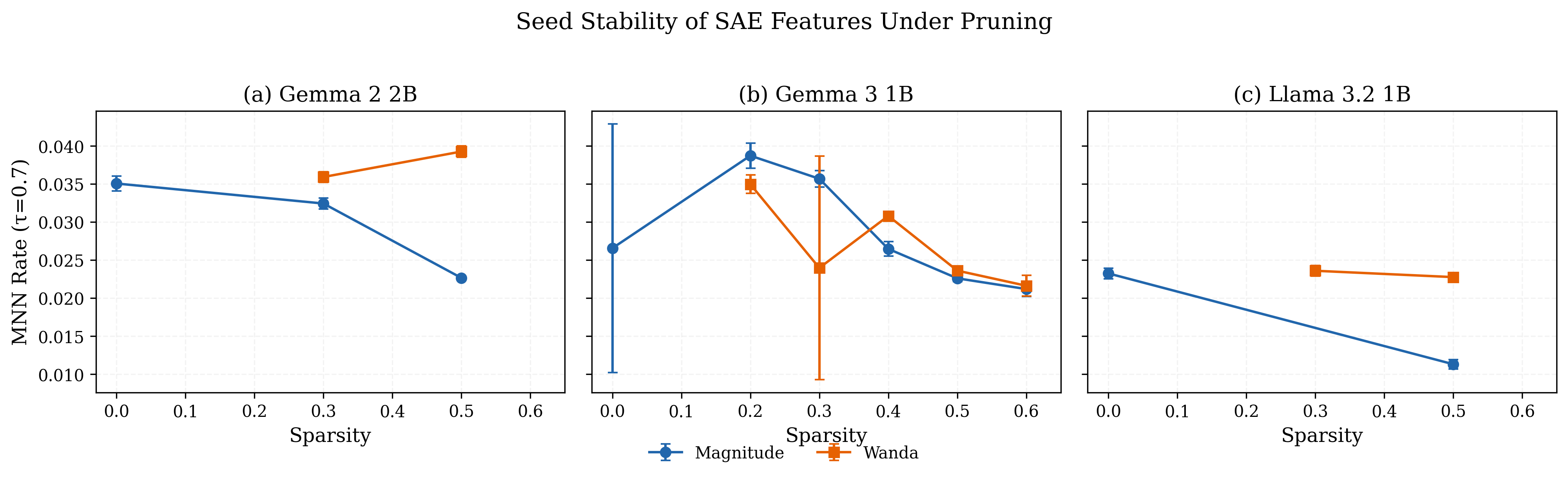}
    \caption{Seed stability (MNN rate at $\tau = 0.7$) as a function of sparsity. Error bars show standard deviation across seed pairs. While absolute MNN rates are low (2--4\%), the \emph{pattern} of degradation with sparsity is consistent.}
    \label{fig:f1}
\end{figure}

SAE features are sensitive to random initialization. At $\tau = 0.7$, the MNN rate between SAEs trained with different seeds on the \emph{same} model ranges from 1.1\% to 3.9\% across all conditions (Figure~\ref{fig:f1}). This means that fewer than 4\% of features trained with one seed can be uniquely matched to features trained with another seed at high cosine similarity.

This low absolute match rate is expected for TopK SAEs with large dictionary sizes ($d_{\text{SAE}} = 8 \times d_{\text{model}}$), where the activation space admits many equivalent sparse decompositions. Critically, however, the \emph{pattern} of MNN degradation with sparsity is highly consistent across seed pairs (standard deviations $<$0.002 in most conditions). For Gemma~3, magnitude pruning shows a monotonic decline from 2.7\% (dense) to 2.1\% (60\% sparsity), while Wanda shows a similar trend from 3.5\% (20\%) to 2.2\% (60\%). Llama shows the sharpest drop under magnitude pruning: from 2.3\% (dense) to 1.1\% (50\% sparsity).

We conclude that while individual SAE features are not reproducible across seeds, the statistical properties of the feature population (and their response to pruning) are stable. All subsequent analyses use seed~0 as the reference for cross-condition comparisons, which is justified by this consistency.

Although absolute seed-to-seed MNN rates are low (mean 2.8\% at 
$\tau = 0.7$), the cross-condition MNN rates we report for 
dense$\to$pruned comparisons (50--80\% at low sparsity) are 
18--29$\times$ above this seed-variability floor. This large 
separation confirms that the structural changes we measure reflect 
genuine effects of pruning, not artifacts of the inherent 
variability in SAE training.

\subsection{RQ2: Feature Survival}
\label{sec:rq2}

\begin{figure}[t]
    \centering
    \includegraphics[width=\textwidth]{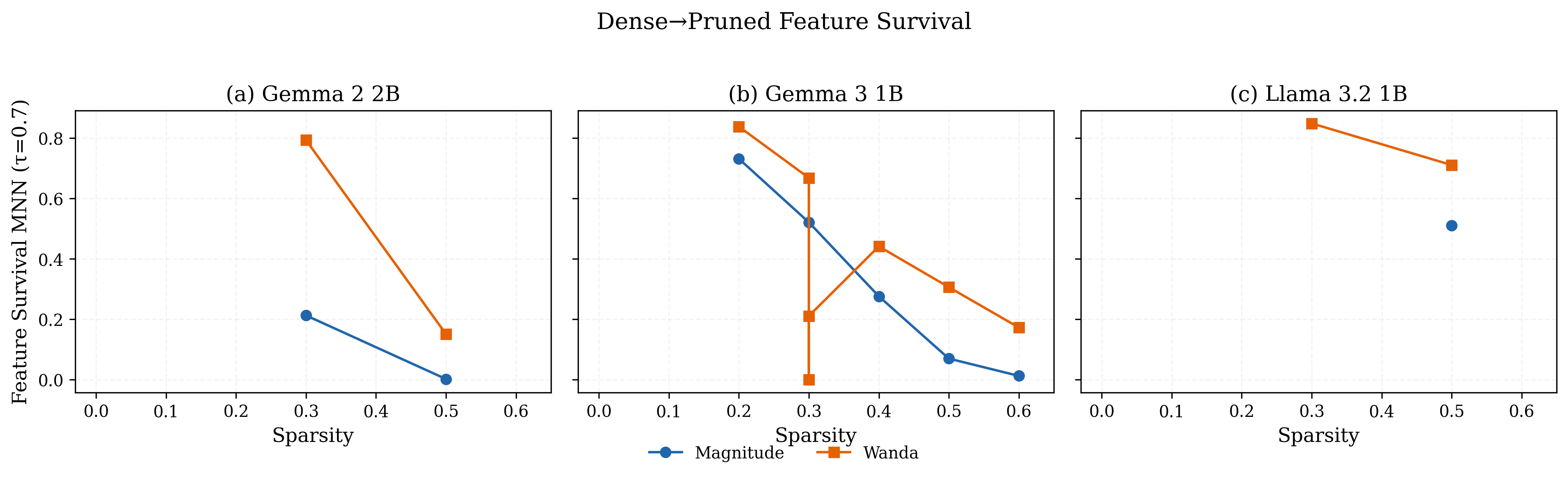}
    \caption{Dense$\to$pruned feature survival (MNN at $\tau = 0.7$). Wanda consistently preserves more features than magnitude pruning across all models and sparsity levels.}
    \label{fig:f2}
\end{figure}

Feature survival measures how many dense-model features can be recovered in the pruned-model SAE dictionary. Figure~\ref{fig:f2} shows the MNN survival rate at $\tau = 0.7$ across models and sparsity levels.

The key finding is that \textbf{Wanda preserves feature structure substantially better than magnitude pruning}. At 30\% sparsity: Gemma~3 shows MNN of 0.669 (Wanda) versus 0.521 (magnitude), a $1.3\times$ advantage. The gap widens dramatically for Gemma~2: 0.794 (Wanda) versus 0.213 (magnitude), a $3.7\times$ difference. For Llama at 50\% sparsity, Wanda achieves 0.711 compared to magnitude's 0.511.

Survival degrades monotonically with sparsity for both methods. For Gemma~3, magnitude pruning survival drops from 0.732 (20\%) to 0.012 (60\%), while Wanda degrades more gracefully from 0.839 (20\%) to 0.173 (60\%). Notably, even at 50\% sparsity---where model perplexity has degraded substantially---Wanda preserves 30.6\% of Gemma~3 features and 71.1\% of Llama features at $\tau = 0.7$.

These results establish that Wanda's activation-aware pruning criterion preserves not only task performance but also the internal feature geometry of the model. The per-row pruning structure of Wanda appears to redistribute weight removal in a way that respects the directions in activation space that the model has learned to use.

\subsection{RQ3: Transferability of Pre-trained SAEs}
\label{sec:rq3}

\begin{figure}[t]
    \centering
    \includegraphics[width=0.75\textwidth]{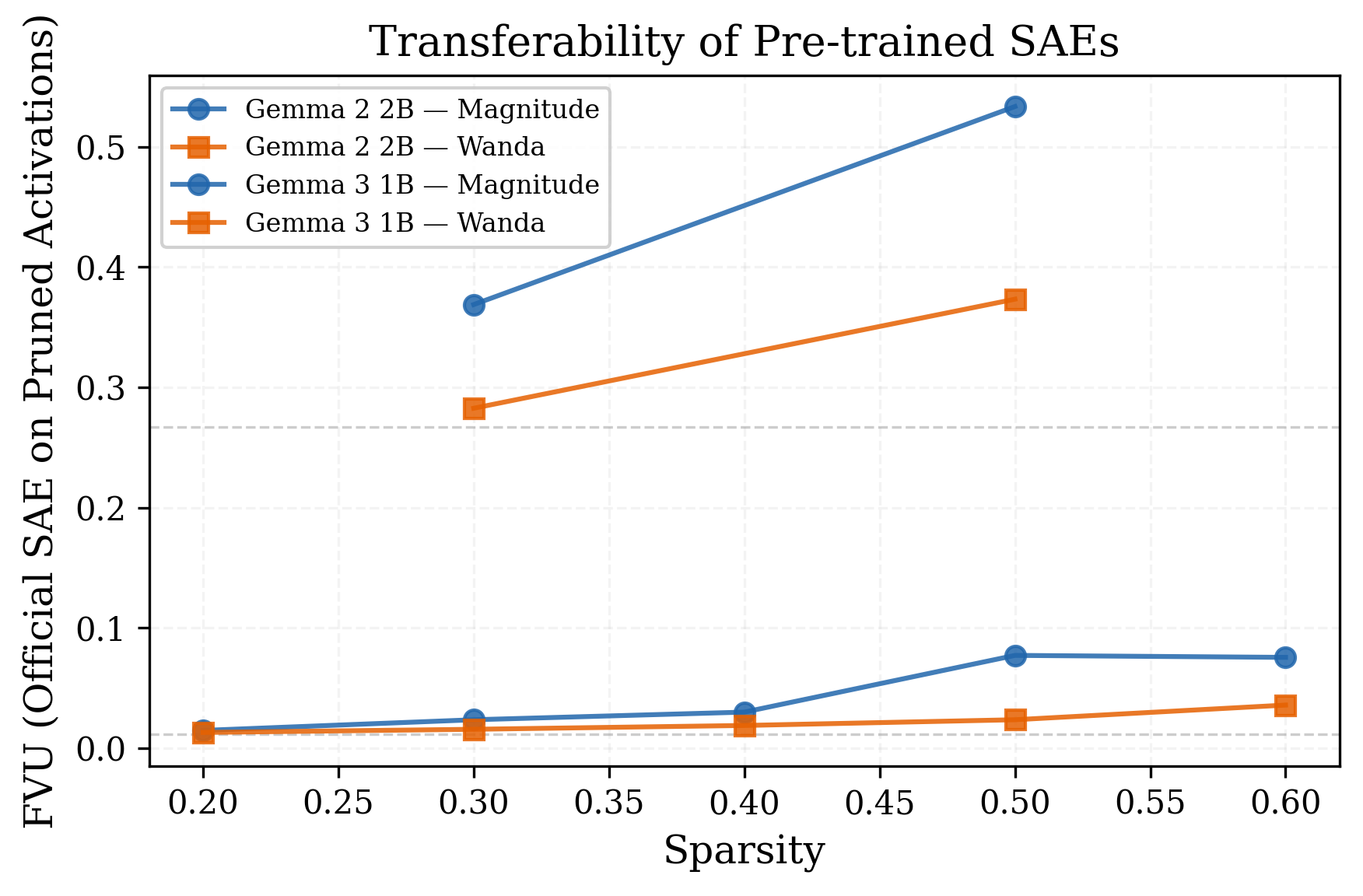}
    \caption{Transferability of pre-trained (official) SAEs to pruned model activations. FVU is plotted against sparsity for Gemma~3 1B and Gemma~2 2B, separated by pruning method. Wanda-pruned activations remain well within the reconstructable range.}
    \label{fig:f3}
\end{figure}

We evaluate whether official Gemma Scope SAEs~\cite{lieberum2024gemma}, trained on dense model activations, can still reconstruct activations from pruned models. Table~\ref{tab:transfer} reports Fraction of Variance Unexplained (FVU) for two models with available official SAEs. Figure~\ref{fig:f3} shows the transferability trends visually.

\begin{table}[t]
\centering
\caption{Transferability: FVU of official (dense-trained) SAEs applied to pruned model activations. Lower FVU indicates better reconstruction. Wanda-pruned activations remain well-reconstructable up to 50\% sparsity.}
\label{tab:transfer}
\small
\begin{tabular}{llrrr}
\toprule
\textbf{Model} & \textbf{Method} & \textbf{Sparsity} & \textbf{FVU} & \textbf{L0} \\
\midrule
\multirow{6}{*}{Gemma 3 1B} & Dense & 0\% & 0.011 & 11.4 \\
                              & Magnitude & 20\% & 0.015 & 10.2 \\
                              & Magnitude & 30\% & 0.024 & 10.0 \\
                              & Magnitude & 50\% & 0.077 & 4.5 \\
                              & Wanda & 30\% & 0.016 & 10.4 \\
                              & Wanda & 50\% & 0.024 & 11.0 \\
\midrule
\multirow{5}{*}{Gemma 2 2B} & Dense & 0\% & 0.267 & 89.4 \\
                              & Magnitude & 30\% & 0.369 & 52.9 \\
                              & Magnitude & 50\% & 0.533 & 57.9 \\
                              & Wanda & 30\% & 0.283 & 93.1 \\
                              & Wanda & 50\% & 0.373 & 111.9 \\
\bottomrule
\end{tabular}
\end{table}

For Gemma~3, the official SAE achieves a remarkable FVU of only 0.024 on Wanda-pruned activations at 50\% sparsity, barely above the dense baseline of 0.011. Magnitude pruning causes larger degradation (FVU = 0.077 at 50\%), with a notable drop in L0 from 11.4 to 4.5, suggesting that many features become inactive. For Gemma~2, Wanda at 30\% sparsity (FVU = 0.283) is much closer to the dense baseline (0.267) than magnitude at the same sparsity (0.369).

The practical implication is significant: \textbf{practitioners who prune with Wanda can continue using existing SAE dictionaries without retraining}, at least up to moderate sparsity levels. This is particularly valuable given the computational cost of training SAEs at scale.

\subsection{RQ4: Feature Fragility by Firing Rate}
\label{sec:rq4}

Our most striking finding concerns the relationship between a feature's firing rate (how often it activates across the evaluation corpus) and its survival under pruning. We bin alive features from the dense SAE into quintiles by firing rate (Q1 = rarest, Q5 = most frequent) and measure the survival rate of each quintile when matching against a pruned SAE.

\begin{figure}[t]
    \centering
    \includegraphics[width=\textwidth]{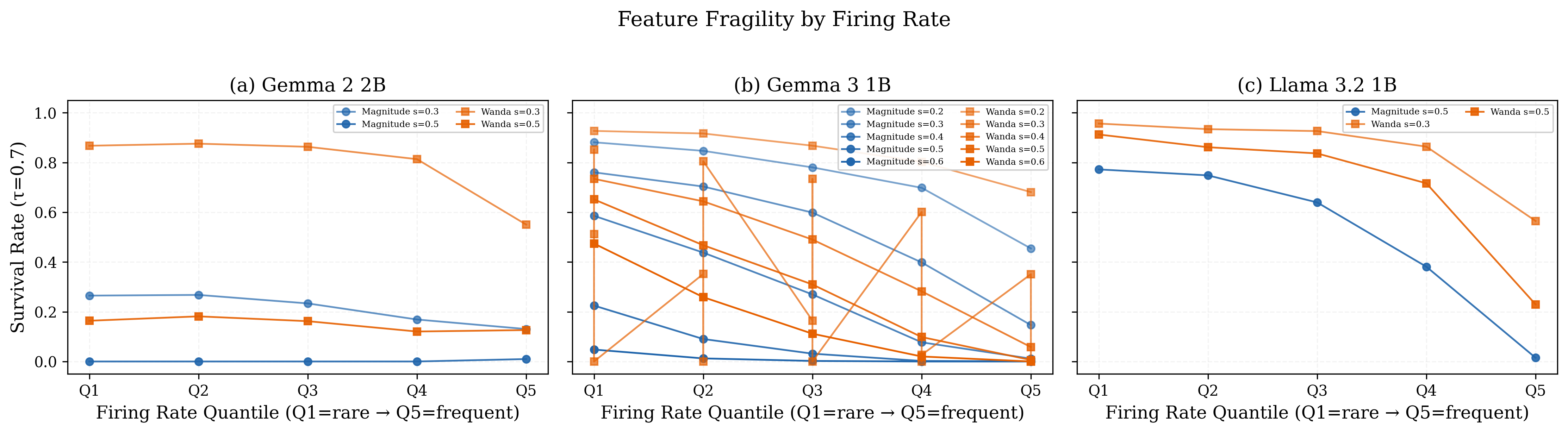}
    \caption{Feature survival rate by firing rate quintile ($\tau = 0.7$). Across all models and pruning methods, rare features (Q1) survive substantially better than frequent ones (Q5). Lines represent different pruning method--sparsity combinations.}
    \label{fig:f4}
\end{figure}

\textbf{Rare features survive pruning far better than frequent ones.} Figure~\ref{fig:f4} shows this pattern across all three model families. Table~\ref{tab:fragility} presents representative Q1-to-Q5 survival ratios.

\begin{table}[t]
\centering
\caption{Feature survival rate by firing rate quintile ($\tau = 0.7$). In all conditions, rare features (Q1) survive at substantially higher rates than frequent features (Q5). The ratio column shows the Q1/Q5 survival advantage.}
\label{tab:fragility}
\small
\begin{tabular}{llrrrr}
\toprule
\textbf{Model} & \textbf{Condition} & \textbf{Q1 (rare)} & \textbf{Q5 (freq)} & \textbf{Ratio} \\
\midrule
\multirow{4}{*}{Gemma 3 1B} & Mag.\ $s=0.3$ & 0.760 & 0.146 & 5.2$\times$ \\
                              & Wanda $s=0.3$ & 0.852 & 0.351 & 2.4$\times$ \\
                              & Wanda $s=0.5$ & 0.652 & 0.006 & 109$\times$ \\
                              & Mag.\ $s=0.5$ & 0.225 & 0.001 & 225$\times$ \\
\midrule
\multirow{2}{*}{Gemma 2 2B} & Wanda $s=0.3$ & 0.867 & 0.550 & 1.6$\times$ \\
                              & Mag.\ $s=0.3$ & 0.265 & 0.130 & 2.0$\times$ \\
\midrule
\multirow{3}{*}{Llama 3.2 1B} & Wanda $s=0.3$ & 0.957 & 0.565 & 1.7$\times$ \\
                                & Wanda $s=0.5$ & 0.913 & 0.229 & 4.0$\times$ \\
                                & Mag.\ $s=0.5$ & 0.772 & 0.015 & 51$\times$ \\
\bottomrule
\end{tabular}
\end{table}

The within-condition Spearman correlation between mean firing rate 
and survival rate is $\rho = -1.0$ (exact permutation test, 
$p = 0.017$, two-tailed) in 11 of 17 experimental conditions, 
and strongly negative ($\rho \leq -0.8$) in 15 of 17. The 
remaining two conditions are edge cases where catastrophic 
pruning destroys nearly all features regardless of firing rate. 
This relationship is robust across all three model families, 
both pruning methods, and all sparsity levels tested.

We hypothesize that this pattern arises because high-firing-rate features encode generic, broadly activated information (e.g., common syntactic patterns, frequent token co-occurrences). These generic features occupy heavily shared weight subspaces and are therefore more vulnerable when weights are removed. In contrast, rare features encode specialized information that relies on specific, less-overlapping weight configurations that pruning is less likely to disrupt simultaneously.

\subsection{Predictive Model: Firing Rate as a Pre-Pruning Diagnostic}
\label{sec:predictive}

\begin{figure}[t]
    \centering
    \includegraphics[width=\textwidth]{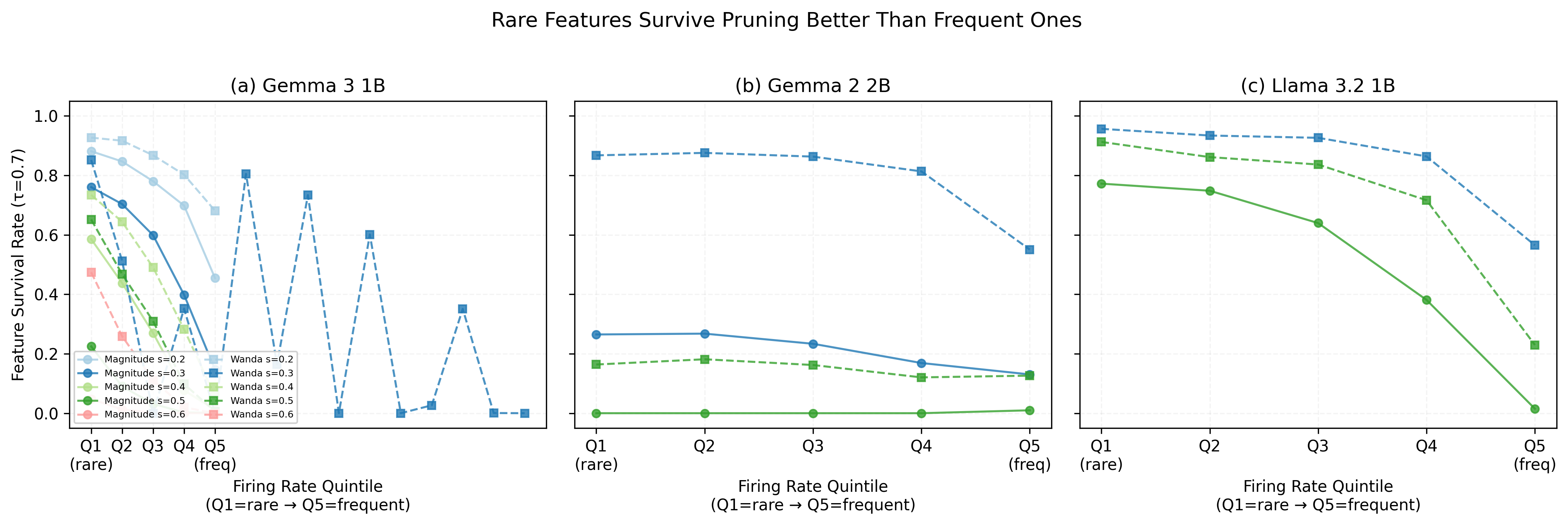}
    \caption{Feature fragility by firing rate, separated by pruning method. Blue: magnitude pruning; orange: Wanda. The monotonic decline from Q1 (rare) to Q5 (frequent) is robust across all models, methods, and sparsity levels.}
    \label{fig:f6}
\end{figure}

The strong monotonic relationship between firing rate and survival motivates a practical diagnostic tool. We fit a logistic regression model to predict whether a feature will survive pruning ($P(\text{survival rate} > 0.5)$) using two predictors: $\log(\text{firing rate})$ and sparsity. Figure~\ref{fig:f6} provides an alternative view of the fragility data, grouped by pruning method.

At $\tau = 0.7$, the model achieves an AUC of 0.796 and accuracy of 69.5\%. At the more lenient threshold $\tau = 0.6$, performance improves to AUC = 0.807 and accuracy of 76.8\%. At $\tau = 0.8$, AUC reaches 0.809 with 83.2\% accuracy. In all cases, the coefficient for $\log(\text{firing rate})$ is negative (ranging from $-1.2$ to $-2.1$), confirming that higher firing rate predicts lower survival probability. The coefficient for sparsity is also consistently negative ($-1.4$ to $-2.0$).

This result provides a practical guideline: \emph{before pruning a language model, practitioners can train an SAE on the dense model and identify features with high firing rates (Q4--Q5 quintiles). These features are the most likely to be destroyed by pruning. Conversely, rare features (Q1--Q2) are robust to pruning up to 40--50\% sparsity, suggesting they encode specialized information that the network actively preserves.}

\subsection{RQ5: Causal Relevance --- An Informative Null Result}
\label{sec:rq5}

A natural hypothesis is that features that survive pruning should be causally more important to the model's output than those that are destroyed. We tested this hypothesis by comparing the mean KL divergence upon ablation of ``robust'' features (high cosine match between dense and pruned SAEs) versus ``fragile'' features (low cosine match).

\begin{figure}[t]
    \centering
    \includegraphics[width=\textwidth]{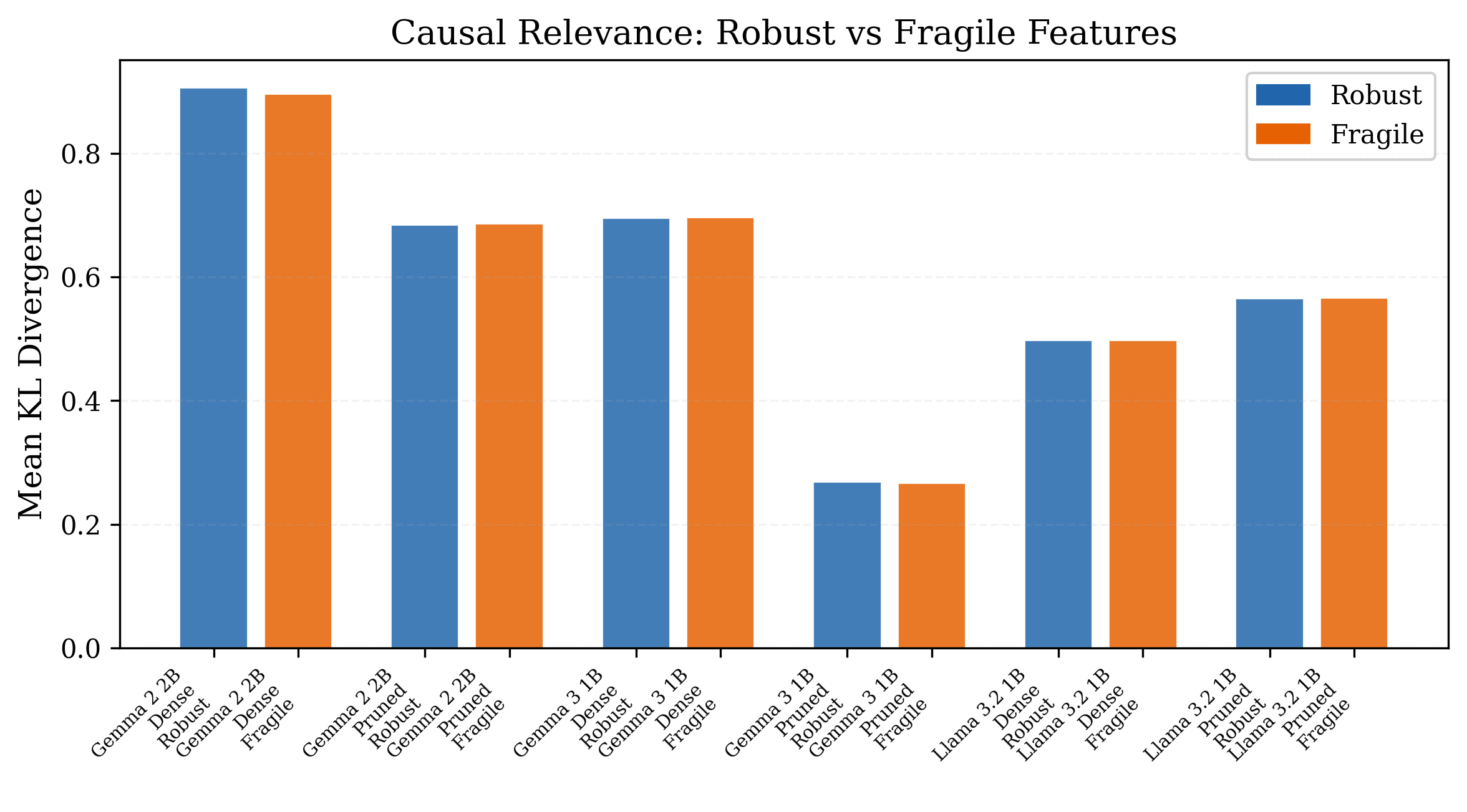}
    \caption{Mean KL divergence upon single-feature ablation for 
robust versus fragile features. Across all models and conditions, 
robust and fragile features show nearly identical causal effects, 
indicating a dissociation between geometric survival and causal 
importance.}
    \label{fig:f5}
\end{figure}

\textbf{The hypothesis is not supported.} Table~\ref{tab:causal} shows that robust and fragile features produce nearly identical KL divergences across all six model--condition pairs.

\begin{table}[t]
\centering
\caption{Mean KL divergence upon single-feature ablation. Robust and fragile features show nearly identical causal effects across all models and conditions. $\Delta$ is the absolute percent difference.}
\label{tab:causal}
\small
\begin{tabular}{llrrr}
\toprule
\textbf{Model} & \textbf{Condition} & \textbf{Robust KL} & \textbf{Fragile KL} & $\boldsymbol{\Delta}$ \\
\midrule
\multirow{2}{*}{Gemma 3 1B} & Dense & 0.695 & 0.696 & 0.1\% \\
                              & Pruned (Wanda 30\%) & 0.268 & 0.266 & 0.6\% \\
\midrule
\multirow{2}{*}{Gemma 2 2B} & Dense & 0.907 & 0.896 & 1.2\% \\
                              & Pruned (Wanda 30\%) & 0.684 & 0.686 & 0.2\% \\
\midrule
\multirow{2}{*}{Llama 3.2 1B} & Dense & 0.497 & 0.498 & 0.1\% \\
                                & Pruned (Wanda 30\%) & 0.565 & 0.566 & 0.2\% \\
\bottomrule
\end{tabular}
\end{table}

The maximum difference between robust and fragile KL is 1.2\% (Gemma~2 dense), with most conditions showing $<$0.5\% difference. This result is consistent across all three model families.

This null result carries two important implications. First, \textbf{geometric feature survival does not predict functional importance}. A feature that maintains high cosine similarity between its dense and pruned decoder vectors is not more causally relevant to the model's outputs. This dissociation suggests that cosine similarity in the decoder space captures something about the \emph{direction} a feature encodes, but not about how much the model \emph{relies} on that direction for its predictions.

Second, we observe that \textbf{individual feature ablations have less effect in pruned models}: mean KL drops from 0.695 to 0.268 (Gemma~3), from 0.907 to 0.684 (Gemma~2), and rises slightly from 0.497 to 0.565 (Llama). The decrease in Gemma models suggests that pruning forces the model to distribute information more broadly across features, making it more robust to individual feature ablations. The slight increase in Llama may reflect a different compensatory strategy.

\section{Discussion}
\label{sec:discussion}

\subsection{Cross-Model Patterns}

Several findings are remarkably consistent across all three model families: (i)~rare features survive better than frequent ones in every condition tested; (ii)~Wanda outperforms magnitude pruning for feature preservation; (iii)~geometric survival does not predict causal importance. These patterns hold despite the models spanning two architectural families (Gemma and Llama), different sizes (1B and 2B parameters), and different training procedures.

The divergence in SAE reconstruction quality (FVU) across model families is intriguing. Gemma~3's decreasing FVU with sparsity suggests that pruning simplifies the activation geometry, making it \emph{easier} for SAEs to reconstruct. This could indicate that pruning removes redundant representational capacity, concentrating activations into a lower-dimensional subspace. In contrast, Llama's increasing FVU suggests that pruning introduces noise or moves activations into regions that are harder to decompose sparsely. Understanding this divergence is an important direction for future work.

\subsection{Connection to the Lottery Ticket Hypothesis}

Our finding that rare features disproportionately survive pruning resonates with the Lottery Ticket Hypothesis~\cite{frankle2019lottery}. If pruning preserves a ``winning ticket'' subnetwork, the features that survive should be those most tightly coupled to this subnetwork. Rare features, which activate on specific, infrequent inputs, may correspond to specialized circuits within the winning ticket. Frequent features, by contrast, may represent generic processing that can be reconstructed from the surviving weights in alternative configurations---explaining why they are ``destroyed'' in the geometric sense (low cosine match) but not in the functional sense (equal causal effect).

\subsection{Practical Recommendations}

Based on our findings, we offer three practical recommendations for practitioners who prune language models and wish to maintain interpretability:

\begin{enumerate}
    \item \textbf{Prefer Wanda over magnitude pruning} when interpretability is a concern. Wanda preserves up to $3.7\times$ more features at matched sparsity and enables reuse of pre-trained SAEs.
    \item \textbf{Pre-trained SAEs remain usable on Wanda-pruned models.} For Gemma models, official Gemma Scope SAEs achieve FVU $<$ 0.025 on Wanda-pruned activations at 50\% sparsity, only $2.1\times$ the dense baseline.
    \item \textbf{Monitor high-firing-rate features before pruning.} Features in the Q4--Q5 firing rate quintiles are the most vulnerable to pruning. If specific interpretable features are important for deployment (e.g., safety-relevant features), check their firing rates---rare features are much more likely to survive.
\end{enumerate}

\section{Limitations}
\label{sec:limitations}

\paragraph{Model Scale.} All models in this study are in the 1--2B parameter range. It remains unclear whether the patterns we observe---particularly the rare-feature survival advantage---hold at larger scales (7B, 70B), where the dynamics of pruning and representation may differ substantially.

\paragraph{Context Length.} We use a context length of 256 tokens for activation extraction, which is shorter than the models' full context window. This likely affects perplexity measurements (the anomalously high Gemma~2 dense PPL of 410 may partly reflect this mismatch) and could influence which features are activated. However, relative differences between conditions are still informative.

\paragraph{Float16 Overflow.} Two runs at layer~18 of Gemma~3 produced NaN activations due to float16 overflow at deep layers, limiting our layer-depth analysis. We report early (L6) and mid (L12) layer results for Gemma~3 and recommend using bfloat16 or float32 for activation extraction at deep layers in future work.

\paragraph{Seed Sensitivity.} The low absolute MNN rates (2--4\%) mean that cross-condition comparisons are made between dictionaries that are themselves highly variable. While the \emph{patterns} are consistent, the specific features identified as ``robust'' or ``fragile'' may differ across seed choices.

\paragraph{Single Pruning Application.} We study one-shot pruning without fine-tuning. Pruning followed by retraining (as in the Lottery Ticket framework) may yield different feature dynamics, as the model has the opportunity to reorganize its representations.

\section{Conclusion}
\label{sec:conclusion}

We have presented the first systematic study of how weight pruning reshapes the internal feature geometry of language models as revealed by Sparse Autoencoders. Across three model families, two pruning methods, and six sparsity levels, we find that:

\begin{enumerate}
    \item Rare SAE features survive pruning far better than frequent ones ($\rho \leq -0.8$ between firing rate and survival in 15/17 conditions), suggesting that pruning acts as implicit feature selection.
    \item Wanda pruning preserves feature structure up to $3.7\times$ better than magnitude pruning, and pre-trained SAEs remain viable on Wanda-pruned activations up to 50\% sparsity.
    \item Geometric feature survival does not predict causal importance: robust and fragile features produce nearly identical effects when ablated ($<$1.2\% difference in all conditions).
    \item A simple logistic model using firing rate and sparsity can predict feature survival with AUC $>$ 0.79, providing a practical pre-pruning diagnostic.
\end{enumerate}

These findings have implications for both the pruning and interpretability communities. For pruning researchers, they suggest that task-level metrics alone may miss important changes in internal structure. For interpretability researchers, they demonstrate that SAE-based analysis tools require careful re-evaluation when applied to compressed models---but also that, with appropriate pruning methods, existing interpretability infrastructure can be largely preserved.

\section*{Data and Code Availability}

All code, data, and trained SAE artifacts are publicly available at \url{https://github.com/hborobia/sae-pruning-paper}. The repository includes: (i)~a fully reproducible Colab notebook that regenerates all experiments from scratch; (ii)~all result tables in CSV format; (iii)~paper-ready figures; and (iv)~the experimental run matrix. Pre-trained SAEs and activation caches are available upon request due to their large size (~50GB).


\appendix
\setcounter{table}{0}
\renewcommand{\thetable}{\Alph{section}.\arabic{table}}
\section{Full Experimental Matrix}
\label{app:matrix}

Table~\ref{tab:full_matrix} summarizes the complete experimental design. Tier A1 runs cover the full sparsity grid on Gemma~3 1B at layer~12. Tier A3 extends to early (L6) and late (L18) layers. Tier B1 covers Gemma~2 2B, and Tier C1 covers Llama~3.2 1B. All runs use residual stream activations (site: \texttt{resid\_post}).

\begin{table}[H]
\centering
\caption{Summary of experimental runs by tier. Seeds column shows number of SAE random seeds per run.}
\label{tab:full_matrix}
\small
\begin{tabular}{llllrrr}
\toprule
\textbf{Tier} & \textbf{Model} & \textbf{Layer} & \textbf{Purpose} & \textbf{Runs} & \textbf{Seeds} & \textbf{Steps} \\
\midrule
A1 & Gemma 3 1B & L12 & Primary grid & 11 & 3--5 & 20{,}000 \\
A3 & Gemma 3 1B & L6, L18 & Layer depth & 4 & 3 & 12{,}000 \\
B1 & Gemma 2 2B & L12 & Anchor model & 5 & 3--5 & 20{,}000 \\
C1 & Llama 3.2 1B & L8 & Cross-family & 4 & 3--5 & 12{,}000 \\
\bottomrule
\end{tabular}
\end{table}

\setcounter{table}{0}
\section{SAE Reconstruction Quality}
\label{app:sae_quality}

Table~\ref{tab:sae_quality} reports mean SAE FVU and alive feature counts across seeds for all main (mid-layer) runs.

\begin{table}[H]
\centering
\caption{SAE reconstruction quality (FVU) averaged over seeds. The large majority of SAEs achieved near-complete feature utilization with our training procedure.}
\label{tab:sae_quality}
\footnotesize    
\begin{tabular}{llrrrr}
\toprule
\textbf{Model} & \textbf{Method} & \textbf{Sparsity} & \textbf{FVU} & \textbf{Alive} & $\boldsymbol{n}$ \\
\midrule
\multirow{11}{*}{Gemma 3 1B}
  & Magnitude & 0.0 & 0.367 & 9{,}216 & 5 \\
  & Magnitude & 0.2 & 0.355 & 9{,}216 & 3 \\
  & Magnitude & 0.3 & 0.335 & 9{,}216 & 3 \\
  & Magnitude & 0.4 & 0.293 & 9{,}216 & 3 \\
  & Magnitude & 0.5 & 0.285 & 9{,}216 & 3 \\
  & Magnitude & 0.6 & 0.351 & 9{,}216 & 3 \\
  & Wanda & 0.2 & 0.363 & 9{,}216 & 3 \\
  & Wanda & 0.3 & 0.350 & 9{,}216 & 5 \\
  & Wanda & 0.4 & 0.308 & 9{,}216 & 3 \\
  & Wanda & 0.5 & 0.278 & 9{,}216 & 3 \\
  & Wanda & 0.6 & 0.241 & 9{,}216 & 3 \\
\midrule
\multirow{5}{*}{Gemma 2 2B}
  & Magnitude & 0.0 & 0.409 & 18{,}432 & 5 \\
  & Magnitude & 0.3 & 0.377 & 18{,}432 & 3 \\
  & Magnitude & 0.5 & 0.439 & 18{,}432 & 3 \\
  & Wanda & 0.3 & 0.402 & 18{,}432 & 3 \\
  & Wanda & 0.5 & 0.379 & 18{,}432 & 3 \\
\midrule
\multirow{4}{*}{Llama 3.2 1B}
  & Magnitude & 0.0 & 0.356 & 16{,}384 & 5 \\
  & Magnitude & 0.5 & 0.487 & 16{,}384 & 3 \\
  & Wanda & 0.3 & 0.373 & 16{,}384 & 3 \\
  & Wanda & 0.5 & 0.376 & 16{,}384 & 3 \\
\bottomrule
\end{tabular}
\end{table}

\end{document}